\documentclass{article}

\usepackage{PRIMEarxiv}

\usepackage[utf8]{inputenc} 
\usepackage[T1]{fontenc}    
\usepackage{url}            
\usepackage{booktabs}       
\usepackage{amsfonts}       
\usepackage{nicefrac}       
\usepackage{microtype}      
\usepackage{lipsum}
\usepackage{fancyhdr}       
\usepackage{graphicx}       
\graphicspath{{media/}}     
\usepackage{amsmath}
\usepackage{amssymb}
\usepackage{graphicx}
\usepackage{subcaption}
\usepackage[bottom]{footmisc}
\usepackage[ruled,vlined]{algorithm2e}
\usepackage{multirow}

\title{NetTraj: A Network-based Vehicle Trajectory Prediction Model with Directional Representation and Spatiotemporal Attention Mechanisms
}

\author{
  Yuebing Liang \\
  Department of Urban Planning and Design \\
  University of Hong Kong \\
  Hong Kong\\
  \texttt{yuebingliang@connect.hku.hk} \\
   \And
  Zhan Zhao\thanks{Corresponding Author} \\
  Department of Urban Planning and Design \\
  University of Hong Kong \\
  Hong Kong\\
  \texttt{zhanzhao@hku.hk} \\
}

\begin{document}
\maketitle

\begin{abstract}
Trajectory prediction of vehicles in city-scale road networks is of great importance to various location-based applications such as vehicle navigation, traffic management, and location-based recommendations. Existing methods typically represent a trajectory as a sequence of grid cells, road segments or intention sets. None of them is ideal, as the cell-based representation ignores the road network structures and the other two are less efficient in analyzing city-scale road networks. Moreover, previous models barely leverage spatial dependencies or only consider them at the grid cell level, ignoring the non-Euclidean spatial structure shaped by irregular road networks. To address these problems, we propose a network-based vehicle trajectory prediction model named NetTraj, which represents each trajectory as a sequence of intersections and associated movement directions, and then feeds them into a LSTM encoder-decoder network for future trajectory generation. Furthermore, we introduce a local graph attention mechanism to capture network-level spatial dependencies of trajectories, and a temporal attention mechanism with a sliding context window to capture both short- and long-term temporal dependencies in trajectory data. Extensive experiments based on two real-world large-scale taxi trajectory datasets show that NetTraj outperforms the existing state-of-the-art methods for vehicle trajectory prediction, validating the effectiveness of the proposed trajectory representation method and spatiotemporal attention mechanisms.
\end{abstract}


\section{Introduction}
With the widespread use of mobile sensors, such as GPS devices and smartphones, it is possible to track all kinds of moving objects in a near real-time fashion. The increasing prevalence of localization technologies has resulted in large volumes of trajectory data which record the spatiotemporal footprints of vehicles. These data often contain valuable information, which gives rise to many location-based services and applications such as vehicle navigation \cite{ziebart_navigate_2008}, traffic management \cite{li_trajectory_2020} and location-based recommendations \cite{kong_time-location-relationship_2017}. Many of these applications rely on the ability to dynamically predict vehicle routing behavior. For example, if the path of vehicles is known in advance, intelligent transportation systems can provide personalized assistance or recommendation functions to drivers, such as dynamic rerouting, personalized risk assessment and mitigation, speed advice and engine management systems. In addition, trajectory prediction is essential for traffic managers to anticipate possible upcoming hazards, future congestion and travel delays, and deploy traffic control and management strategies accordingly. Based on trajectory prediction, it's also possible to develop customized location-based advertising for those most likely to pass through certain business outlets and shops \cite{rathore2019scalable}. 

Extensive researches have been done to address the problem of vehicle trajectory prediction in previous works. Traditional approaches mostly depend on probabilistic graphical models for vehicle trajectory prediction \cite{simmons_learning_2006, asahara_pedestrian-movement_2011, gambs_next_2012}, which usually rely on one or two previous locations to predict the next one based on variants of Markov models. Recently, with the rapid advance in deep learning, a number of deep neural network models have been developed to address vehicle behavior prediction problems \cite{park_sequence--sequence_2018, wang_multi-vehicle_2020}. Many related approaches are developed for autonomous driving applications \cite{mozaffari_deep_2020}. They usually focus on a specific road section and predict the states of the nearby vehicles in the next few seconds. These models are not applicable to the city-scale vehicle trajectory prediction problem due to the drastically different time and spatial scales. Only a few recent studies attempted to predict the route choice of a vehicle approaching an intersection by pre-defining an intention set of going straight, turning left, turning right and so on \cite{phillips_generalizable_2017,ding_predicting_2019}. Despite extensive researches, there still exist a few research gaps to be addressed:
\begin{itemize}
    \item Existing methods represent vehicle trajectories as sequences of grid cells, road segments or intention sets, none of which are ideal. First, grid-cell based representation fails to consider the topological structure of the road network, and is not suitable for applications based on the road network (e.g., dynamic routing). Second, the use of road segments may lead to data sparsity and high dimensionality issues, especially when the road network is large and complex. Third, for the intention-based trajectory representation, it is hard to predetermine all possible driving intentions in a big and complex driving network \cite{mozaffari_deep_2020}.
    \item Previous models barely leverage spatial dependencies or only consider them at the grid cell level, ignoring the non-Euclidean spatial structure shaped by irregular road networks.. Although Graph Neural Networks (GNNs) have shown effectiveness in modeling graph structures, it restricts the input as a single graph and is not applicable to trajectory data. A new way to model the network-level spatial dependencies of trajectory data is needed.
    \item Existing approaches mostly focus on the next position prediction and are difficult to generalize for long-term sequential patterns. One main reason is that it is computationally costly for sequence prediction since the number of candidate sequences is exponential to the length of the sequence \cite{qiao_predicting_2018}. As a result, existing trajectory prediction algorithms tend to have relatively low prediction accuracies for long-range trajectory prediction.
\end{itemize}

Therefore, despite existing studies, vehicle trajectory prediction in city-scale road networks is still a challenging problem and requires innovative solutions. To solve this problem, this paper proposes a network-based sequence-to-sequence (seq2seq) model with directional representation and spatiotemporal attention mechanisms, named NetTraj, for vehicle trajectory prediction in city-scale road networks. In the proposed model, a direction-based trajectory representation method is performed so that both the input and output sequences of road segments are replaced with sequences of intersections and associated movement directions. Subsequently, an embedding layer transforms the intersection and direction information into a latent space and a local graph attention layer is used to capture spatial dependencies in the road network. The embedded information is then fed to a LSTM encoder-decoder module with a sliding temporal attention mechanism to capture temporal interaction across the input and output trajectories. Finally, an output layer is used to generate future trajectories. The model is evaluated based on two real-world datasets of taxi GPS trajectories from Shanghai and Beijing. The main contributions of this paper are as follows:
\begin{itemize}
    \item We propose a seq2seq model for vehicle trajectory prediction in a city-scale road network. To the best of our knowledge, our proposed model is the first deep learning-based trajectory prediction model that accounts for the structure of road networks at such large scale. 
    \item We introduce a direction-based trajectory representation approach which represents each road segment as a pair of intersection and associated movement direction. This helps reduce the prediction dimensionality from over 100,000 road segments to a small number of direction labels (8 in our experiment settings), which improves both the modeling efficiency and prediction accuracy.
    \item We develop a local graph attention mechanism to model dynamic spatial dependencies of trajectory data in road networks, and a temporal attention mechanism with a sliding context window to model both short- and long-term temporal dependencies in trajectories.
    \item Comprehensive experiments are conducted based on real-world taxi trajectory data from Shanghai and Beijing. The results show that our model outperforms state-of-the-art methods, and all the new model components improve the prediction performance. 
\end{itemize}

This paper is structured as follows. Section \ref{Related Work} discusses related work and limitations. The problem statement and our proposed NetTraj model is introduced in Section \ref{Methodology}, and our numerical experiments, including datasets and results are presented in Section \ref{Experiments}. The paper is concluded in Section \ref{Conclusion}.

\section{Related Work}\label{Related Work}
As prior studies adopted different representations of trajectories, we will first review the methods for trajectory representation, before discussing specific models for trajectory prediction. 

\subsection{Trajectory Representation}
Trajectory representation is a mathematical abstraction of trajectory data necessary for modeling. Depending on how a vehicle’s location is represented, there are three major trajectory representation methods: cell-based, link-based, and intention-based. 
In cell-based representation, the geographic area is divided into regular-shaped cells, and the vehicle’s location is aggregated to the corresponding cell. Zhou et al. \cite{zhou_x201csemi-lazyx201d_2013} presented a “semi-lazy” approach for trajectory prediction which divides the study area into a set of uniform cells and the model outputs the predicted grid cell that the vehicle will visit next. Pecher et al. \cite{pecher_data-driven_2016} proposed a vehicle trajectory prediction model based on Markov models and artificial neural networks by dividing the study area to a grid of cells, each a 1.25km-by-1.25km area. A k-d tree-based space discretization was performed in \cite{ebel_destination_2020}, aggregating GPS locations to discrete cells, each of which contains a defined number of data points. 
While a cell-based representation is useful for certain applications (e.g., destination prediction), this method has several limitations for trajectory prediction: (1) it does not consider the topological structure of the actual road network; and (2) aggregation by land cells is typically too coarse for certain applications (e.g., routing) that require the precise vehicle path along the road network. 

Link-based representation maps the vehicles’ GPS locations to road network segments using map matching techniques \cite{lou_map-matching_2009}. The trajectory is expressed as a series of road segments, and the problem of trajectory prediction is to determine the next series of road segments the vehicle will visit. Liu and Karimi \cite{liu_location_2006} introduced a probability-based model that estimates the probability of taking each road segment at each intersection. In \cite{ziebart_navigate_2008}, the road network is represented with trajectory-level characteristics and taxi trajectories are predicted given known destinations based on probabilistic reasoning from observed context-aware behavior. Tang et al. \cite{tang_inferring_2018} adopted a Hidden Markov Model (HMM) to predict the next route choice of taxis and employed a linear motion function to estimate the taxi’s future location. The main drawback of link-based representation is the large number of road segments in big cities, which not only increases the data sparsity problem \cite{xue2015solving}, but also poses challenges to statistical modeling. 

Intention-based representation describes vehicle behavior with an intention set such as going straight, turning left, turning right and so on. Philips et al. \cite{phillips_generalizable_2017} applied an LSTM model to predict whether a driver will turn left, turn right or go straight at an intersection. 
Ding et al. \cite{ding_predicting_2019} extended the intention-based representation to highway driving scenarios, and predicted lane change and lane keeping behavior of vehicles. One main drawback of these approaches is that it is difficult to predetermine all possible driving intentions in large-scale complex road networks \cite{mozaffari_deep_2020}.

\subsection{Trajectory Prediction}
Existing trajectory prediction methods can be grouped into two categories: probabilistic graphical models and deep learning models. Probabilistic graphical models, especially variants of Markov-chain (MC) models, have been widely used to mine frequent patterns in mobility sequences. Asahara et al. \cite{asahara_pedestrian-movement_2011} proposed a mixed MC model that considers both individual and generic effects 
for pedestrian movement prediction. An extended MC model was adopted to predict the next location of an individual based on previous $n$ locations \cite{gambs_next_2012}. More generally, a Bayesian $n$-gram model was introduced to mine spatiotemporal dependencies for next trip prediction \cite{zhao_individual_2018}. 
In addition, HMM is also often used to model mobility sequences or vehicle trajectories \cite{simmons_learning_2006, Mo_individual_2021}. 
There are three main drawbacks in these methods: (1) they rely on one or two previous locations for trajectory prediction while fail to consider long-term temporal dependencies; (2) they mostly focus on the next location prediction while perform poorly for long-range trajectory prediction; and (3) they are not as flexible for capturing the environment and other external factors. 

Compared with Markov models, deep neural networks are generally better at modeling spatiotemporal dependencies in sequential data \cite{wang2020deep}. Most deep learning-based models employ Recurrent Neural Networks (RNN) and its variants for trajectory prediction. Liu et al. \cite{liu2016predicting} proposed a next location prediction model by feeding temporal and spatial contexts to the recurrent layer. 
Park et al. \cite{park_sequence--sequence_2018} applied a LSTM encoder-decoder to predict the next locations of vehicle trajectories on an occupancy grid map. To capture long-term temporal dependencies, an attention-based recurrent architecture was proposed in \cite{bahdanau_neural_2016}, which adaptively pays more attention to relevant hidden states. Inspired by this, Li et al. \cite{li_hierarchical_2020} developed a hierarchical temporal attention-based recurrent model for human mobility prediction. In \cite{capobianco_deep_2021}, an attention mechanism was used to enhance the recurrent layer for vessel trajectory prediction. In addition to RNNs, Convolutional Neural Networks (CNN) are also used to model mobility sequences for trajectory prediction \cite{tang2018personalized, chen2020cem}.  
Recently, a model named Transformer is proposed in \cite{vaswani_attention_2017}, which relies entirely on self-attention networks for sequence modeling. Based upon this, a self-attention based model is proposed for sequential prediction in \cite{kang2018self}. Compared with RNNs and CNNs, self-attention networks are good at capturing long-term temporal dependencies with less computation.

Although the above studies demonstrate the effectiveness of deep learning models in modeling sequential data, they barely consider spatial dependencies. To leverage spatial information, some recent studies exploit spatial intervals to represent pairwise correlations between locations. Huang et al. \cite{huang2019attention} used travel distance between consecutive visits to represent spatial features. A relation matrix was defined in \cite{luo2021stan} based on spatial distance to capture spatial information of all locations. Another group of studies devide the study area into uniform grids and extract spatial dependencies between grid cells. Lian et al. \cite{lian2020geography} used hierarchical griddings for spatial discretization and used self-attention layers to leverage geographic information. Lv et al. \cite{lv_t-conv_2018} mapped trajectories to two-dimensional grid images and applied Convolutional Neural Networks (CNN) for trajectory destination prediction. 
However, the non-Euclidean spatial dependencies in irregular road networks are not considered. Recently, GNNs have shown effectiveness to capture network-level spatial dependencies in traffic studies \cite{peng2020spatial, peng2021dynamic}. By combining GNNs with attention mechanisms, Graph Attention Networks \cite{velickovic_graph_2018} enable learning dynamic weights between neighborhood nodes without priori knowledge. However, GNNs cannot be directly used in the trajectory prediction problem, as the input of GNNs has to be a single graph and it is infeasible to directly feed trajectories to GNNs. Despite progress in deep learning for trajectory prediction, existing approaches are usually implemented in a small study area with cell-based trajectory representation. They are not applicable to large-scale road networks, as they fail to capture the non-Eucliean spatial dependencies in irregular road networks and the large number of output neurons can limit the model efficiency and prediction performance.

\section{Methodology}\label{Methodology}
\begin{figure*}[h]
  \centering
  \includegraphics[width=0.85\linewidth]{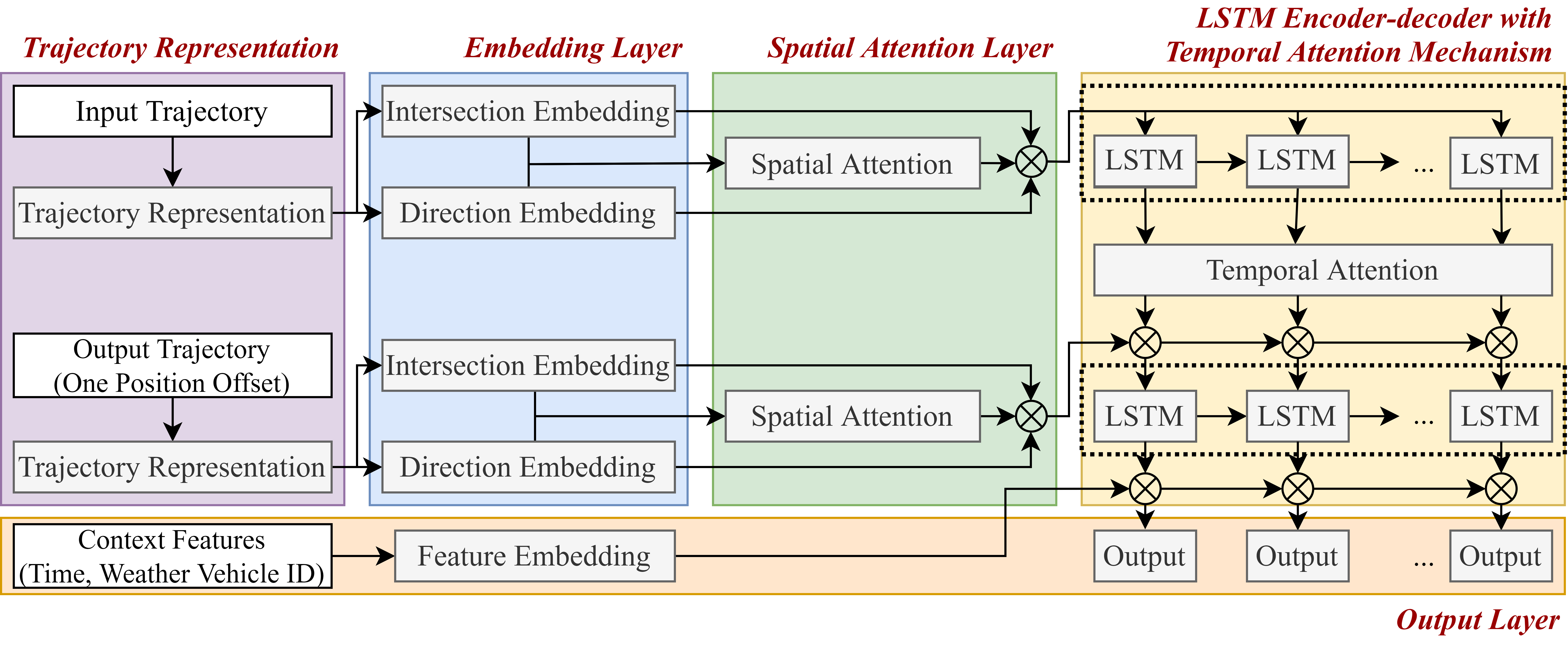}
  \caption{Architecture of the proposed trajectory prediction model ($\bigotimes$ represents the concatenation operation)}
  \label{fig:architecture}
\end{figure*}

In this section, we first introduce our problem statement and then propose a new framework for trajectory prediction in city-scale road networks. As shown in Fig. \ref{fig:architecture}, the proposed framework comprises of five modules: (1) a trajectory representation module that transforms the input and output trajectories into an intersection sequence and a direction sequence, (2) an embedding layer that encodes the intersection and direction information, (3) a spatial attention layer that captures the spatial dependencies in urban road networks, (4) a LSTM encoder-decoder module with temporal attention mechanism that captures temporal dependencies in trajectories, and (5) an output layer that incorporates other context features and outputs the predicted trajectory.

\subsection{Definitions and Problem Statement}
\textit{Definition 1 (Road Network)}: The road network is represented as a directed graph  $G=(N,E)$, which is composed of a set of intersection nodes $N$ and a set of road segments $E$. Each road segment $e \in E$ can be represented as $e=(e.source,e.target)$, where $e.source,e.target \in N$ are the source and target nodes of road segment $e$ respectively. 

\textit{Definition 2 (Trajectory)}: a trajectory $T$ of length $l$ comprises $l$ connected road segments in time series, and is expressed as $T= \{e_1, e_2, ...e_l\}$  where $e_i\in E,1 \leq i \leq l$, $e_i.target=e_{i+1}.source$.  

\textit{Problem}: given a prefix trajectory of length $l_{in}$ denoted as $X = \{e_{1-l_{in}}, ...e_{-1}, e_0\}$ where a vehicle orderly visited, the trajectory prediction problem aims to predict the future trajectory of length $l_{out}$ denoted as $Y = \{e_1, e_2, ...e_{l_{out}}\}$ that the vehicle will traverse. 

\subsection{Trajectory Representation}
In order to preserve the road network typology while overcoming the problem of high prediction dimensionality, a direction-based trajectory representation approach is proposed. The general idea of the approach is to map each road segment to a unique label as $e=(e.source,e.direction)$, where $e.direction$ is a categorical feature indicating the rough direction of $e$. As we will demonstrate later, this new representation can greatly simplify the trajectory prediction problem. The direction-based trajectory representation is composed of 3 steps.

\textit{Step 1}: Calculate the specific heading direction of each road segment. Given $e$, we use $e.heading$ to denote the heading direction from $e.source$ to $e.target$. The basis of heading direction is north.

\begin{figure}[h]
  \centering
  \includegraphics[width=0.6\linewidth]{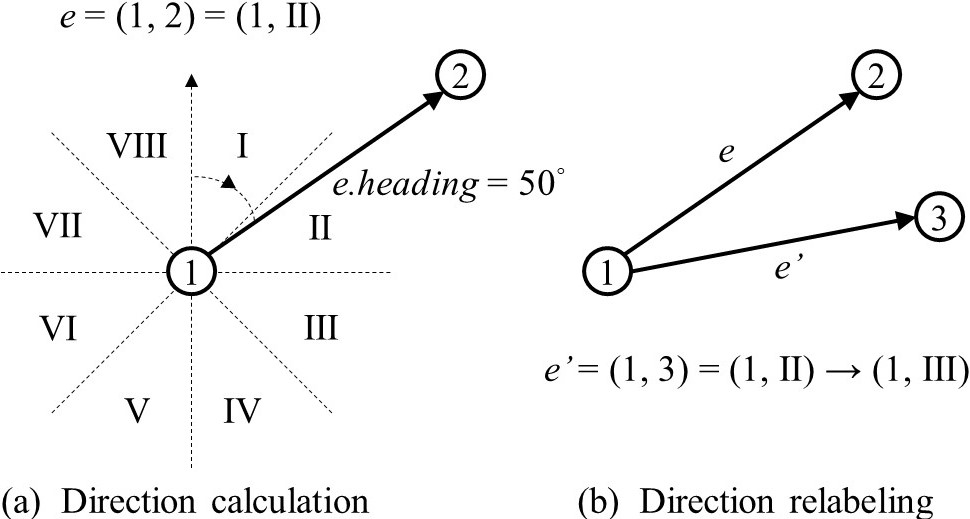}
  \caption{Direction-based trajectory representation approach}
  \label{fig:direct}
\end{figure}

\textit{Step 2}: Discretize $e.heading$ based on predefined intervals. Specifically, the heading range [0, 360) is partitioned into $K$ equal intervals, each having the same width $W = 360 /K$. Each road segment $e$ is mapped to the interval $e.interval$ using:
\begin{equation}
e.interval =  \lfloor \frac{e.heading}{W} \rfloor
\end{equation}
Fig. \ref{fig:direct}a illustrates Steps 1 and 2, assuming $K=8$. In this example, $e_1.heading = 50$ degrees, and $e_1.interval = $II (the second interval).

\textit{Step 3}: Map each road segment $e=(e.source, e.target)$ to a new representation $e=(e.source, e.direction)$. Initially, $e.direction=e.interval$. However, there may be more than one road segments with the same $e.interval$ in some occasions. If road segment $e'$ shares the same representation with another road segment $e$, where $e.source = e'.source$ and $e.interval = e'.interval$, we would redefine $e'.direction$ to the closest \textit{available} interval to $e'.heading$. An interval is available if there is no other road segment starting from the same source with the same interval. Fig.~\ref{fig:direct}b shows an example. Note that while such revision might influence the accuracy of the direction representation of some roads, the influence is trivial since only a small portion of roads require revisions. It is uncommon to see more than one road segments that start from the same intersection and are less than 45 degrees apart. For example, in the road network of Shanghai, when $K=8$ only 2\% road segments require revision to their direction labels.

\begin{figure}[h]
  \centering
  \includegraphics[width=0.6\linewidth]{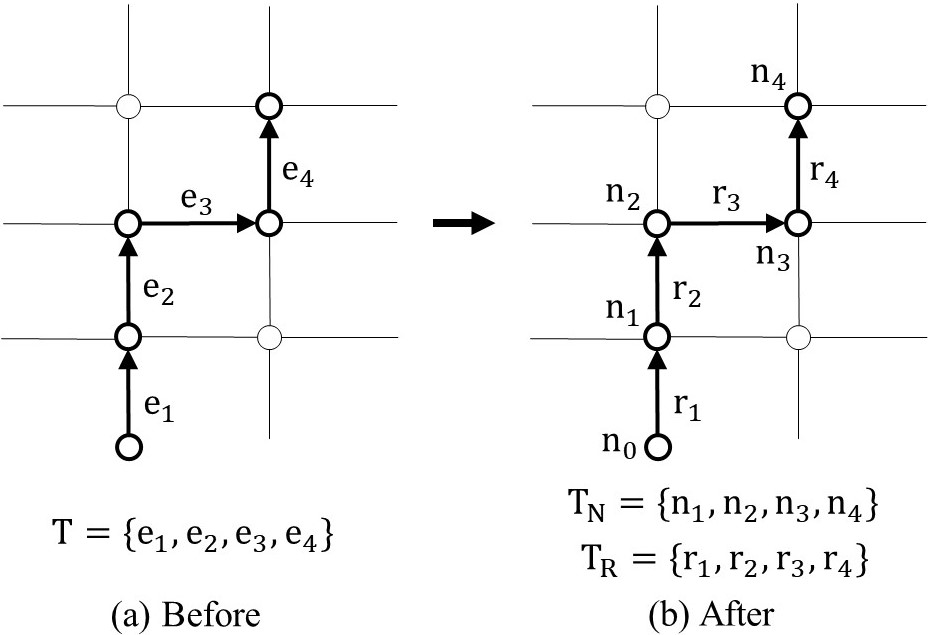}
  \caption{Trajectory before and after direction-based trajectory representation}
  \label{fig:direct_step4}
\end{figure}

Using our proposed direction-based trajectory representation method, each road segment $e$ is mapped to a unique combination $(e.source, e.direction)$. This means that given the source node and direction, we can uniquely identify the target node. For the $i$-th road segment in a trajectory $e_i$, we denote $n_{i-1}=e_i.source$, $n_i=e_i.target$, and $r_i=e_i.direction$. An example of the new trajectory after trajectory representation is shown in Fig. \ref{fig:direct_step4}. Based on this new representation, we can predict a future trajectory $Y = \{e_1,e_2..., e_{l_{out}}\}$ by simply predicting its associated direction sequence $Y_R = \{r_1,r_2...,  r_{l_{out}}\}$ (assuming $n_0$ is given). Algorithm~\ref{alg:alg1} shows how we may use $Y_R$ to reconstruct the corresponding node sequence $Y_N = \{n_1,n_2...,  n_{l_{out}}\}$ as well as $Y$. Note that $Y_R$ has much lower dimensionality than $Y$, and should be more predictable, which is the main advantage of the proposed trajectory presentation method. For the input trajectory $X = \{e_{1-l_{in}}..., e_{-1}, e_0\}$, we would preserve as much information as possible by using both the direction sequence $X_R = \{r_{1-l_{in}}..., r_{-1}, r_0\}$, and node sequence $X_N = \{n_{1-l_{in}}..., n_{-1}, n_0\}$.


\begin{algorithm}[ht]
\SetAlgoLined
\SetKwInOut{KwIn}{Input}
\SetKwInOut{KwOut}{Output}
\KwIn{output direction sequence $Y_R = \{r_1,r_2...,  r_{l_{out}}\}$, the last intersection of input intersection sequence ${n_0}$}
\KwOut{output trajectory $Y = \{e_1,e_2...,  e_{l_{out}}\}$}
 $Y=\{\}$\;
 \For{$i$ in $\{1, 2,... l_{out}\}$}{
  Find the next node $n_i$ according to the unique label $(n_{i-1},r_i)$\;
  $e_i = (n_{i-1}, n_i)$\;
  $Y = Y + \{e_i\}$\;
 }
 \textbf{return} $Y$\;
 \caption{Generating output trajectory $Y$ from output direction sequence $Y_R$}\label{alg:alg1}
\end{algorithm}

\subsection{Embedding Layer}
The node and direction sequences cannot be fed into neural networks directly. The embedding layer maps them to a low-dimensional latent space. Specifically, an intersection embedding layer first maps each intersection node $n\in N $ to a latent vector $v_n\in \mathbb{R}^M$ by learning a parameter matrix $W_n \in \mathbb{R}^{M\times|N|}$. Similarly, the direction embedding layer maps each direction $r$ to a vector $v_r\in \mathbb{R}^D$  by multiplying a parameter matrix $W_r \in \mathbb{R}^{D\times K}$ to its one-hot representation. $M$ and $D$ are the latent vector dimensions for node $n$ and direction $r$ respectively. 
Parameters of $W_n,W_r$ are learned simultaneously with all other parameters through model training.

\subsection{Spatial Attention Layer}
\begin{figure}[h]
  \centering
  \includegraphics[width=0.6\linewidth]{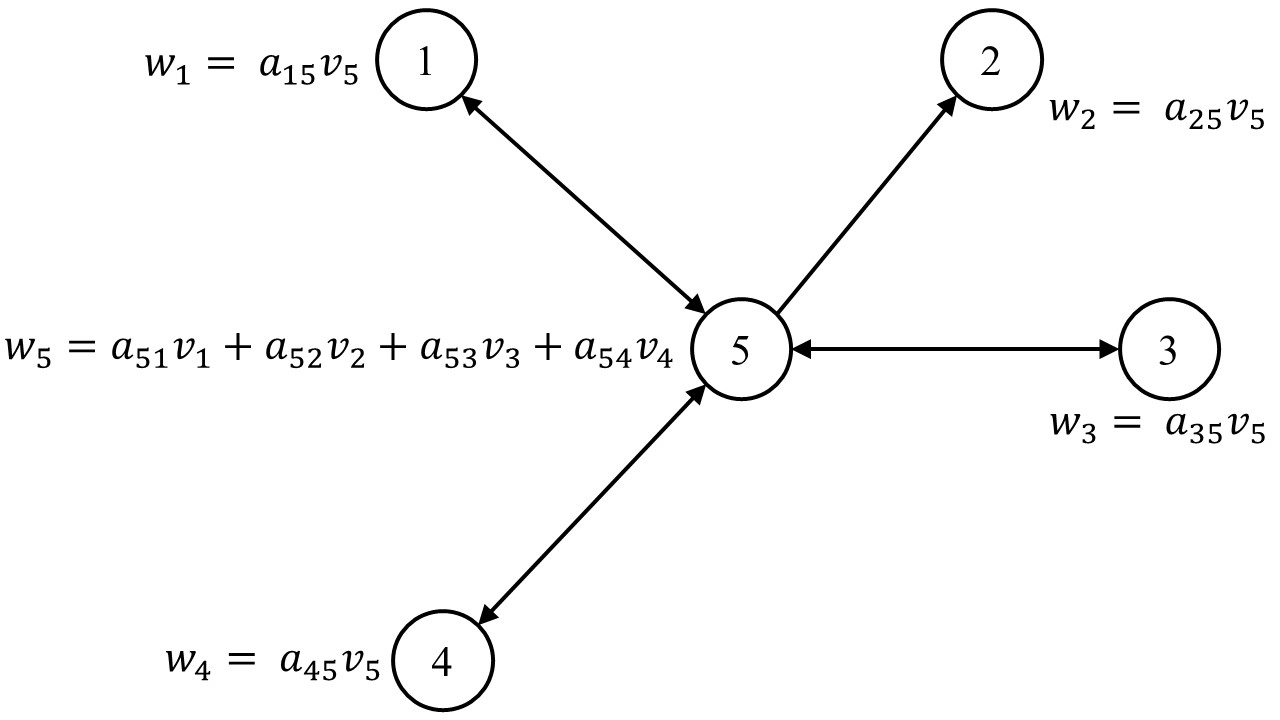}
  \caption{Spatial attention: for each node $n$ in trajectories, a spatial context vector $w_n$ is computed as a weighted sum of the features of its adjacent nodes in the road network.}
  \label{fig:spatial_attn}
\end{figure}

While the embedding layer captures spatial information of each intersection through observed vehicle trajectories, it does not explicitly consider the road network typology. Although GNNs have shown effectiveness in capturing network-level spatial dependencies, it is not applicable to the trajectory prediction problem as the input trajectory cannot be fed to GNNs directly. To overcome the issue, we develop a local graph attention mechanism to model the spatial dependencies of trajectory data. Specifically, for each intersection in the input trajectory, a small graph is formed over the node and its adjacent nodes. Adjacent nodes pass messages to each other and an attention mechanism is applied to calculate the weighted influence (i.e., attention weights) between adjacent nodes. For each intersection node in trajectories, the spatial attention layer generates a context vector as a weighted sum of the features of its adjacent intersections. Fig. \ref{fig:spatial_attn} illustrates the message passing mechanism in the spatial attention layer.

For each intersection $n$ in trajectories, a spatial context vector $w_n \in \mathbb{R}^M$ is built with the following four steps:

\textit{Step 1}: Identify adjacent intersection nodes for an intersection $n$, denoted as $n.neighbor$. We impose a spatial context window where nodes only pass messages to adjacent nodes that connect with node $n$ directly. The spatial context window allows the spatial attention layer to make use of the connectivity structure of the road network and implement message passing more efficiently. 

\textit{Step 2}: Compute the importance score of each adjacent node $n'$ to node $n$ given as:
\begin{equation}
s_{nn'} = [v_n; v_{n'}; v_r]W_s
\end{equation}
where $s_{nn'}\in \mathbb{R}$ is the importance score, $v_n, v_{n'}\in \mathbb{R}^M$ are the embedded vectors of nodes $n$ and $n'$, and $v_r \in \mathbb{R}^D$ is the embedded vector for the direction $r$ associated with node $n$. $v_n, v_{n'}$ and $v_r$ are all obtained from the embedding layer. $W_s \in \mathbb{R}^{(2M+D) \times 1}$ is the weight matrix of a shared linear transformation applied to every node. Note that we incorporate direction features in the importance score computation so that vehicles passing the same intersection from different directions may result in different importance scores. The incorporation of direction features can capture the dynamic spatial dependencies under different scenarios.

\textit{Step 3}: Normalize the importance score using softmax function:
\begin{equation}
a_{nn'}=\frac{\exp(s_{nn'})}{\sum_{n'' \in n.neighbor} \exp(s_{nn''})}
\end{equation}
where $a_{nn'} \in \mathbb{R}$ is the attention weight of adjacent node $n'$ to current node $n$. 

\textit{Step 4}: Compute the context vector $w_n$ as a weighted sum of the features of neighborhood nodes and attention weights:
\begin{equation}
w_n=\sum_{n' \in n.neighbor} a_{nn'}v_{n'}
\end{equation}
where $v_{n'} \in \mathbb{R}^M$ is the embedded vector of node $n'$ obtained from the embedding layer. 

\subsection{LSTM Encoder-decoder with Temporal Attention Mechanisms}
Seq2seq models typically have an encoder-decoder structure. The encoder reads and summarizes the input sequence to an internal representation. The decoder then uses the internal representation to generate an output sequence one element at a time. 
To capture the temporal dependencies within trajectories, we adopt the RNN-based encoder-decoder architecture, which is the state-of-the-art architecture for trajectory prediction problems \cite{park_sequence--sequence_2018,capobianco_deep_2021}. Through empirical comparisons, we choose LSTM as the basic recurrent unit, which achieves relatively good performance in the specific application. In this paper, we represent LSTM as:

\begin{equation}
\label{eq:lstm}
h_t,c_t = LSTM(x_t, h_{t-1}, c_{t-1})
\end{equation}
where $h_t, c_t \in \mathbb{R}^{B}$ are cell memory state vector and hidden state vector respectively, $B$ is the dimension of hidden state vector.


\subsubsection{Encoder}
The input of the encoder is the input node sequence $X_N$ and direction sequence $X_R$. The encoder consists of $S$ stacked LSTM layers. At each time step, given $n_i$ and $r_i$, we learn the embedded vectors $v_{n_i}$ and $v_{r_i}$ from the embedding layer, and then the context vector $w_{n_i}$ from the spatial attention layer. The input vector $x_i$ of the recurrent unit is thus defined as the concatenation of the generated representations, i.e., $x_i=[v_{n_i}; v_{r_i}; w_{n_i}]$, where $x_i \in \mathbb{R}^{2M+D}$. Then, the encoder recursively takes $x_i$ as the input vector and hidden state vector $h_{i-1}$ from previous recurrent unit to update vectors $h_i, c_i$ at each time step through Eq.~\eqref{eq:lstm}.
After $l_{in}$ time steps, the encoder summarizes the whole input sequence into the final vectors $c_0, h_0$ and an output sequence of hidden state vectors $H = \{h_{1-l_{in}}..., h_{-1}, h_0\}$.

\subsubsection{Decoder}
The input of the decoder is the output node sequence $Y_N$ and direction sequence $Y_R$. Note that the output elements are offset by one position to make sure the prediction for position $i$ only depends on the previously known outputs. The decoder also consists of $S$ stacked LSTM layers. Similar to the encoder, given $n_i$ and $r_i$, we learn the embedded vectors $v_{n_i}$ and $v_{r_i}$ from the embedding layer, and then the spatial context vector $w_{n_i} \in \mathbb{R}^B$ from the spatial attention layer. In addition, a temporal context vector $u_i$ is learned from a temporal attention layer that will be introduced later. The input vector for the decoder is $x'_i=[v_{n_i}; v_{r_i}; w_{n_i}; u_i]$, where $x'_i \in \mathbb{R}^{2M+D+B}$. The decoder uses $h_0$ and $c_0$ passed from the encoder as its initial hidden state vector and cell memory state vector. By feeding $x'_i, h_{i-1},c_{i-1}$ to the recurrent unit recursively using Eq.~\eqref{eq:lstm}, the output sequence of hidden state vectors $H'=\{h_1, h_2, ..., h_{l_{out}}\}$ is generated. 

\subsubsection{Sliding Temporal Attention Layer}
\begin{figure}[h]
  \centering
  \includegraphics[width=0.9\linewidth]{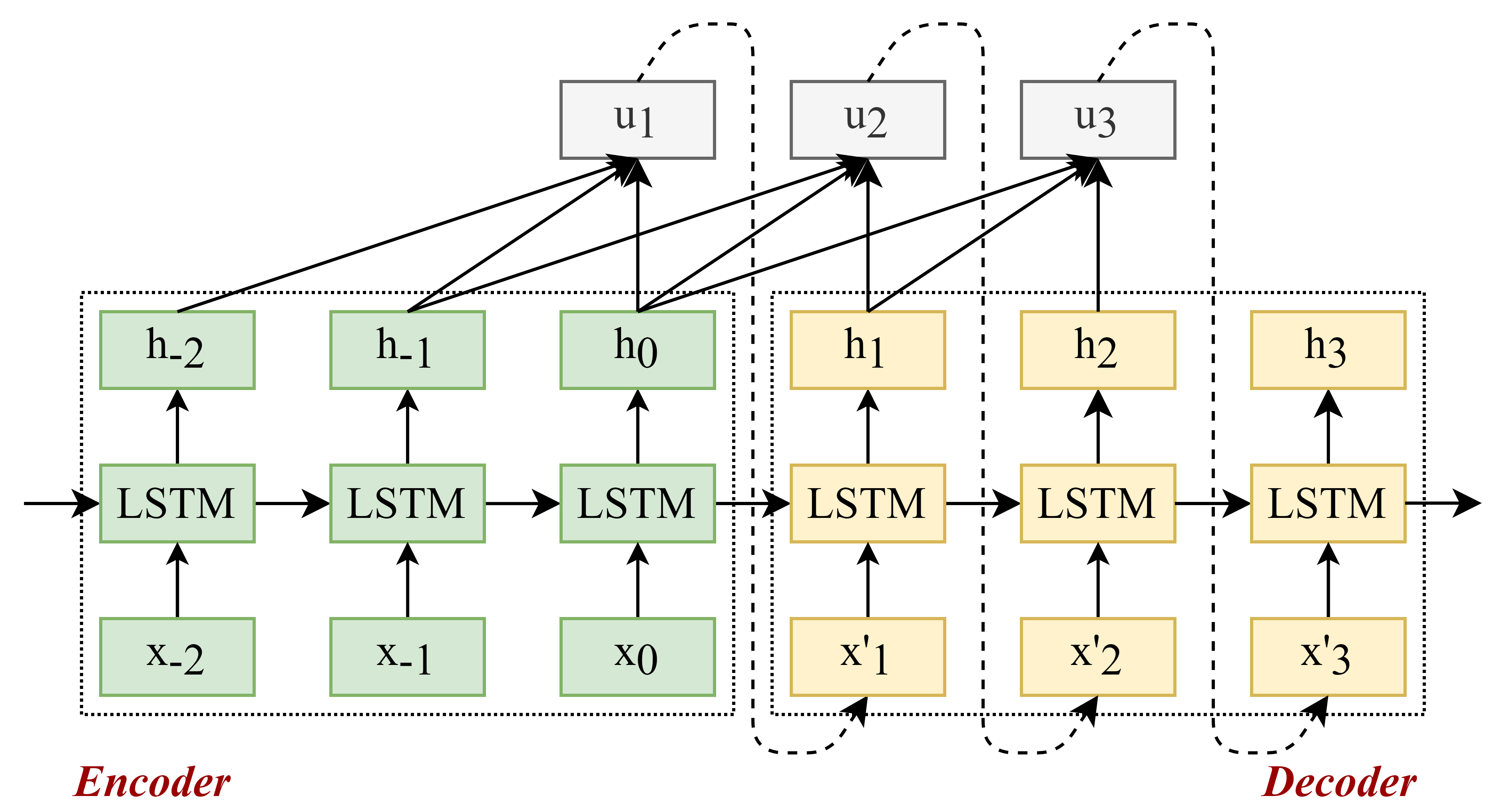}
  \caption{Sliding temporal attention: for each position $i$ in the decoder, a context window $[i-l_{in}, i)$ is used to compute a temporal context vector $u_i$, which is a weighted sum of the hidden states in the window.} 
  \label{fig:temporal_attn}
\end{figure}

Temporal attention mechanisms can capture long-term temporal dependencies by adaptively paying more attention to relevant hidden states in the encoder output sequence $H$ during future sequence generation. It has shown effectiveness in applications such as machine translation \cite{bahdanau_neural_2016} and human mobility prediction \cite{li_hierarchical_2020}. Vehicle trajectory prediction is different from previous applications, because the prediction of the next path relies highly on current state. To capture such temporal dependencies in the temporal attention mechanism, we define a sliding temporal context window $[i-l_{in}, i)$ for each time step $i$, and the sequence $H$ is updated as $\{h_{i-l_{in}},... h_{-1}, h_0, h_1, h_2,... h_{i-1}\}$. A weighted sum of hidden state vectors in the updated sequence $u_i \in \mathbb{R}^B$ is generated to capture temporal dependencies using:
\begin{gather}
u_i = \sum_{j=i-l_{in}}^{i-1} z_{ij}h_j \\
z_{ij} = \frac{\exp(s_{ij})}{\sum_{m=i-l_{in}}^{i-1} \exp(s_{im})} \\
s_{ij}=[h_{i-1}; c_{i-1}; h_j]W_t
\label{eq:temp}
\end{gather}
where $W_t \in \mathbb{R}^{(1+2S)\times B}$ is the weight matrix for attention weight computation. Fig. \ref{fig:temporal_attn} illustrates the mechanism of the sliding temporal attention layer. 


\subsection{Output Layer}
The output layer predicts the subsequent direction sequence given the decoder output sequence $H'$ from the LSTM encoder-decoder. We also incorporate context features which may also influence drivers’ route decisions in the prediction, including time, individual preferences and inclement weather. In our work, the continuous timestamp is divided into $7 \times 24 = 168$ intervals which represents each hour in a week. In addition, we represent weather information as a categorical variable (sunny/cloudy/small rain/rain/heavy rain), and individual information as a unique driver or vehicle identifier representing the inherent routing preferences of each driver. The context features are first mapped to continuous representations using the same embedding method as the embedding layer and then concatenated with the decoder output sequence $H'$. Finally, a simple linear layer is adopted to map the concatenated representation to the dimension of directions $K$ and a softmax function is applied to estimate the probability of each direction.
\begin{equation}
Y_{out} = softmax([H'; V_{time}; V_{weather}; V_{individual}]W_c)
\end{equation}
where $V_{time} \in \mathbb{R}^{l_{out} \times Q_{time}}$, $V_{weather} \in \mathbb{R}^{l_{out} \times Q_{weather}}$, and $V_{individual} \in \mathbb{R}^{l_{out} \times Q_{individual}}$ are the embedded vectors for time, weather and individual information, and $W_c \in \mathbb{R}^{(B+Q_{time}+Q_{weather}+Q_{individual}) \times K}$  is the parameter matrix for linear transformation.  

\subsection{Model Setup}
\subsubsection{Loss function}
the prediction of future directions is essentially to classify which direction the vehicle will next visit. Following studies of multi-class classification, we use cross entropy as the loss function to train the model.
\begin{equation}
loss(\Theta) = -\sum_{i=1}^{l_{out}} \sum_{r=1}^{K} {o_i^r\ln{\hat{P}(o_i^r)}}
\end{equation}
where $\Theta$ denotes the set of parameters in our model, $o_i^r \in \mathbb{R}$ is a dummy variable indicating whether $r$ is the target direction at the $i$-th position in target trajectory and $\hat{P}(o_i^r)$ is its corresponding probability predicted by the model. 

\subsubsection{Scheduled Sampling}
the input of the decoder during training is the output trajectories with one position offset so that the decoder generates predictions given previously known ground truth observations. In model testing, ground truth observations are replaced by predictions generated by the model itself. Specifically, the most likely direction is selected as the predicted direction and the predicted intersection node is obtained using algorithm \ref{alg:alg1}. To mitigate the discrepancy between the input distributions of training and testing, we incorporate scheduled sampling \cite{bengio_scheduled_2015} in the model. During the training process, the model is fed with either the ground truth observation with probability $\alpha$ or the predicted result with probability $1-\alpha$, and $\alpha$ gradually decreases to 0 as the iteration increases.

\section{Experiments}\label{Experiments}
In this section, we evaluate our proposed model by conducting extensive experiments on two real-world vehicle trajectory datasets, and compare it against other state-of-the-art methods.

\subsection{Datasets}
\begin{figure*}[h]
  \centering
  \includegraphics[width=\linewidth]{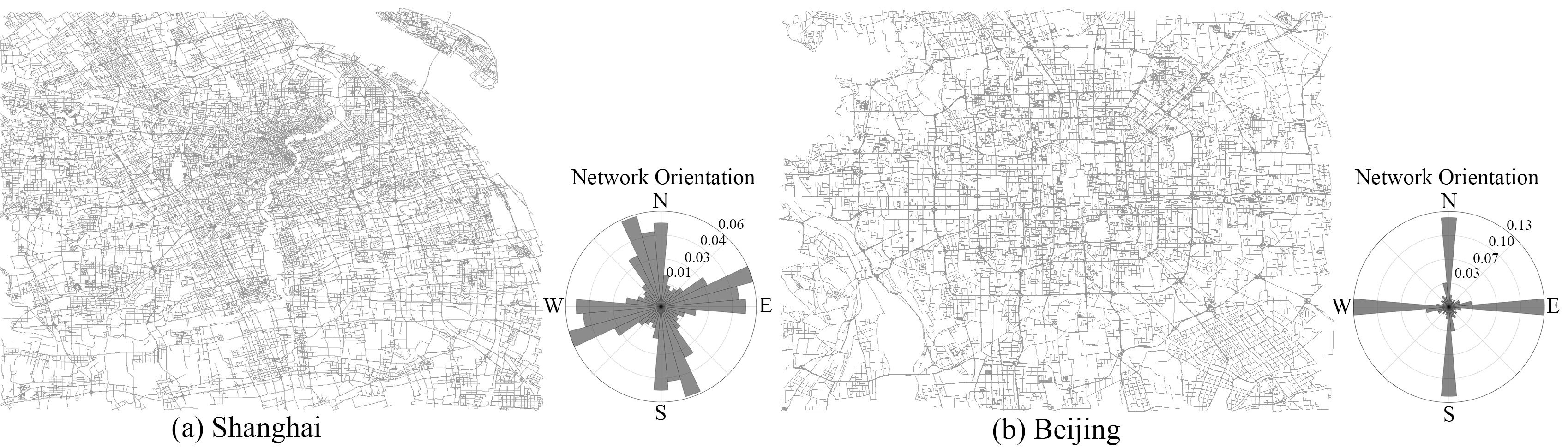}
  \caption{Road networks used in our trajectory prediction experiments}
  \label{fig:dataset}
\end{figure*}

\textit{Shanghai Taxi Trajectory Data (Shanghai Dataset)}: 
The dataset is retrieved from one of the major taxi companies in Shanghai which contains the GPS traces of 7579 taxis from 2015-04-15 to 2015-04-21 in Shanghai, China. The average sampling rate is roughly every 10 seconds, and there are over 4 million records in the dataset. Each record consists of the information of taxi ID, date, time, longitude, latitude and occupied flag, which is a binary indicator flagging whether the taxi is occupied. In our experiment, we take trajectories within a bounding box: [30.85, 31.42, 121.08, 121.85]. The road network is obtained from OpenStreetMap using an open-source tool OSMnx \cite{boeing_osmnx_2017}. The road network consists of 42,990 intersection nodes and 105,426 road segments as shown in Fig. \ref{fig:dataset}a.

\textit{{Beijing Taxi Trajectory Data} (Beijing Dataset)\cite{yuan_t-drive_2010}}: 
This dataset is obtained from the T-Drive project which contains one-week trajectories of 10,357 taxis from 2008-02-02 to 2008-02-08 in Beijing, China. The average sampling rate is 177 seconds and the total number of records is about 15 million. Each record consists of the information of date, time, longitude and latitude. Similarly, we take trajectories within a bounding box: [39.74, 40.05, 116.14, 116.60]. The road network is also obtained from OpenStreetMap, consisting of 30,383 intersection nodes and 69,849 road segments as shown in Fig. \ref{fig:dataset}b.

In addition to taxi GPS data and road network data, we collect weather data from Weather Underground Website. The dataset provides weather condition (e.g., fair, cloudy, rain, storm) every half an hour for Shanghai Data and every 2-3 hours for Beijing Data.

\subsection{Data Preprocessing}
For Shanghai Dataset, we adopt our proposed data preprocessing approach introduced in a previous paper \cite{liang_mining_2021} to extract valid taxi trips from the raw data. After data preprocessing, the possible states of taxis are assigned to the trajectories and invalid trips during which taxis are off-shift or taking a short break are filtered. As most vehicles on the road are en route to their destinations, we exclude cruising taxi trips in the following experiments because their routing patterns are quite different. Beijing Dataset does not contain an occupied flag and is generally more sparse, so we cannot segment taxi trajectories based on the occupancy status. Instead, we simply split the trajectory of each driver to multiple sub-trajectories of 50 records with an average time length of 2.5h. To obtain the trajectories as a sequence of road segments, we use an open-source map matching framework called Fast Map Matching \cite{yang_fast_2018} which provides an implementation of the ST-Match map mapping approach \cite{lou_map-matching_2009}. The training, validation and test sets are split randomly for both datasets. Table \ref{table:dataset} shows the training, validation and test size of both datasets after data preprocessing. To create input and output sequences $X$ and $Y$, we adopt a segmentation method commonly employed in machine translation \cite{vaswani_attention_2017} and time series analysis \cite{capobianco_deep_2021}. In the first step, we concatenate the input-output trajectories in a single sequence and divide it into batches of size $C$ by trimming off remaining tokens, which allows more efficient batch processing. The second step is to use a sliding window of size $Z$ to reduce the data to input sequences of length $l_{in}$ and output sequences of length $l_{out}$. For instance, for a training set containing $L$ trajectories with lengths of $\{l_1, l_2..., l_{L}\}$, the training set will be divided into $C$ sequences of length $l_s = \lfloor \frac {\sum_j^{L} l_j} {C} \rfloor$. For each sequence $T$ at the $i$-th prediction step, the input trajectory is $T_{z \times i : z \times i+l_{in}}$ and the output trajectory is $T_{z \times i+l_{in} : z \times i+l_{in}+l_{out}}$.
\begin{table}[ht]
  \centering \footnotesize
  \caption{Training, validation and test set description}
    \begin{tabular}{c c c}
    \hline
    Dataset & SH & BJ \\
    \hline
    Total Trajectories (Avg. Length) & 520,890 (65) & 149,105 (255) \\
    Training Set (Avg. Length) & 500,890 (65) &	139,105 (255) \\
    Validation Set (Avg. Length) & 10,000 (65) & 5,000 (259) \\
    Test Set (Avg. Length) & 10,000 (65) & 5,000 (251) \\
    \hline
    \end{tabular}%
  \label{table:dataset}%
\end{table}%

\subsection{Evaluation Metrics}
In our experiments, we evaluate the performance of our vehicle trajectory prediction method using the following metrics:
\begin{itemize}
    \item \textit{Distance Error (DE)}: the average edit distance between predicted and actual trajectories. Edit distance has been commonly used to quantify the similarity between two sequences by counting the minimum number of operations required to transform one sequence to the other. The formula of DE is defined as:
    \begin{equation}
    DE = \frac{1}{L'l_{out}} \sum_{i=1}^{L'} {Edit(\hat{Y}_i, Y_i)}
    \end{equation}
    where $L'$ is the number of trajectories in the test set, $\hat{Y}_i, Y_i$ are the $i$th predicted and true trajectories respectively and $Edit(\hat{Y}_i, Y_i)$ is the edit distance between the two. 
    \item \textit{Average Match Ratio (AMR)}: the average ratio of the number of correctly predicted road segments to the length of predicted trajectories. The formula of MR is defined as:
    \begin{equation}
    AMR = \frac{1}{L'l_{out}} \sum_{i=1}^{L'} {\sum_{j=1}^{l_{out}} Match(\hat{e}_{j}^i, e_{j}^i)}
    \end{equation}
    where $\hat{e}_{j}^i, e_{j}^i$ are the predicted and true road segment in the $j$-th position of the $i$-th trajectory. $Match(\hat{e}_{j}^i, e_{j}^i)=1$ if they are the same and 0 otherwise.
    \item \textit{Match Ratio(k) (MR(k))}: the ratio of the number of trajectories with at least $k$ road segments correctly predicted to the number of total trajectories $L'$, where $k=1,..., l_{out}$. For example, MR($l_{out}$) refers to the likelihood of correctly predicting every road segment in a trajectory. The formula of $MR(k)$ is defined as:
    \begin{equation}
    MR(k) = \frac{1}{L'} \sum_{i=1}^{L'} {Match_k(\hat{Y}_i, Y_i)}
    \end{equation}
    where $Match_k(\hat{Y}_i, Y_i)=1$ if $\sum_{j=1}^{l_{out}} Match(\hat{e}_{j}^i, e_{j}^i) \geq k$ and 0 otherwise. By definition, $Match_k(\hat{Y}_i, Y_i)$ decreases with $k$.
\end{itemize}

\subsection{Settings}
All experiments are conducted on a NVIDIA 1080 Ti GPU. We use the stochastic gradient descent optimizer with the batch size of 20, the sliding window size of 5 and the dropout rate of 0.1. The initial learning rate is set to 0.5, and it decays at a rate of 0.8 every epoch. Through intensive empirical optimization, we determine the hyperparameters of our proposed model as:
\begin{itemize}
    \item The number of directions $K=8$
    \item The embedding dimensions of node and direction sequences $M=D=256$
    \item The hidden dimension of the LSTM encoder-decoder network $B=512$
    \item The depth of LSTM stack $S=2$
    \item The embedding dimensions of context features $Q_{time}=Q_{weather}=Q_{individual}=32$
    \item The length of an input trajectory $l_{in}=10$
    \item The length of an output trajectory $l_{out}=5$
\end{itemize}

\subsection{Baseline Models}
As introduced in Section \ref{Related Work}, existing vehicle trajectory prediction mostly adopted probabilistic graphical models and there is no existing deep learning-based method for vehicle trajectory prediction in a city-scale road network. Therefore, we implement several state-of-the-art trajectory prediction models which were originally developed for other applications (e.g., location prediction, POI recommendation), in addition to a basic Markov chain model, for benchmarking.
\begin{itemize}
    \item \textit{Markov Chain (MC)}. We use a first-order Markov Chain, or MC(1), as a benchmark method, which can be extended to predict a sequence of road segments iteratively by conditioning on the previous known or predicted road segments.
    \item \textit{LSTM Encoder-decoder (LSTM)}. A LSTM encoder-decoder architecture was adopted by \cite{park_sequence--sequence_2018} for vehicle trajectory prediction on an occupancy grid map. For comparison, a similar model is implemented for vehicle trajectory prediction in road networks. The encoder takes an input intersection node sequence and the decoder generates a future node sequence.
    \item \textit{Convolutional Sequence Embedding (Caser)}. A CNN-based model was introduced in \cite{tang2018personalized} for next POI recommendation. We implement a similar model for vehicle trajectory prediction, which maps the input trajectory to an embedding matrix and uses multi-dimensional convolutional filters to uncover sequential patterns.
    \item \textit{Attentional LSTM Encoder-decoder (AT-LSTM)}. Capobianco et al. \cite{capobianco_deep_2021} proposed a vessel trajectory prediction model based on LSTM encoder-decoder and applied an attention mechanism to aggregate encoder outputs. We implement a similar model which inputs the latest intersection node sequence and generates a future node sequence for comparison.
    \item \textit{Attentional Spatiotemporal LSTM (ATST-LSTM)}. An attention-based LSTM network was devised in \cite{huang2019attention} for next POI recommendation, which uses spatial and temporal intervals between consecutive positions as contextual information to improve model performance. We adapt the model to an encoder-decoder architecture for fair comparison.
    \item \textit{Self-attention based Sequential Model (SASRec)}. Kang et al. \cite{kang2018self} devised a general sequential recommendation model which relies entirely on self-attention networks dispensing with RNNs and CNNs. A similar self-attention based model is implemented for comparison. 
    \item \textit{Geography-aware Self-attention Network (GeoSAN)}. A geography-aware location prediction model is introduced in \cite{lian2020geography} based on self-attention networks. We implement a similar model for vehicle trajectory prediction. Specifically, the GPS locations of intersection nodes are mapped to hierarchical grids and the spatial information within trajectories is uncovered using a self-attention based geography encoder. 
\end{itemize}

\subsection{Prediction Performance}
\begin{figure}[h]
  \centering
  \includegraphics[width=0.7\linewidth]{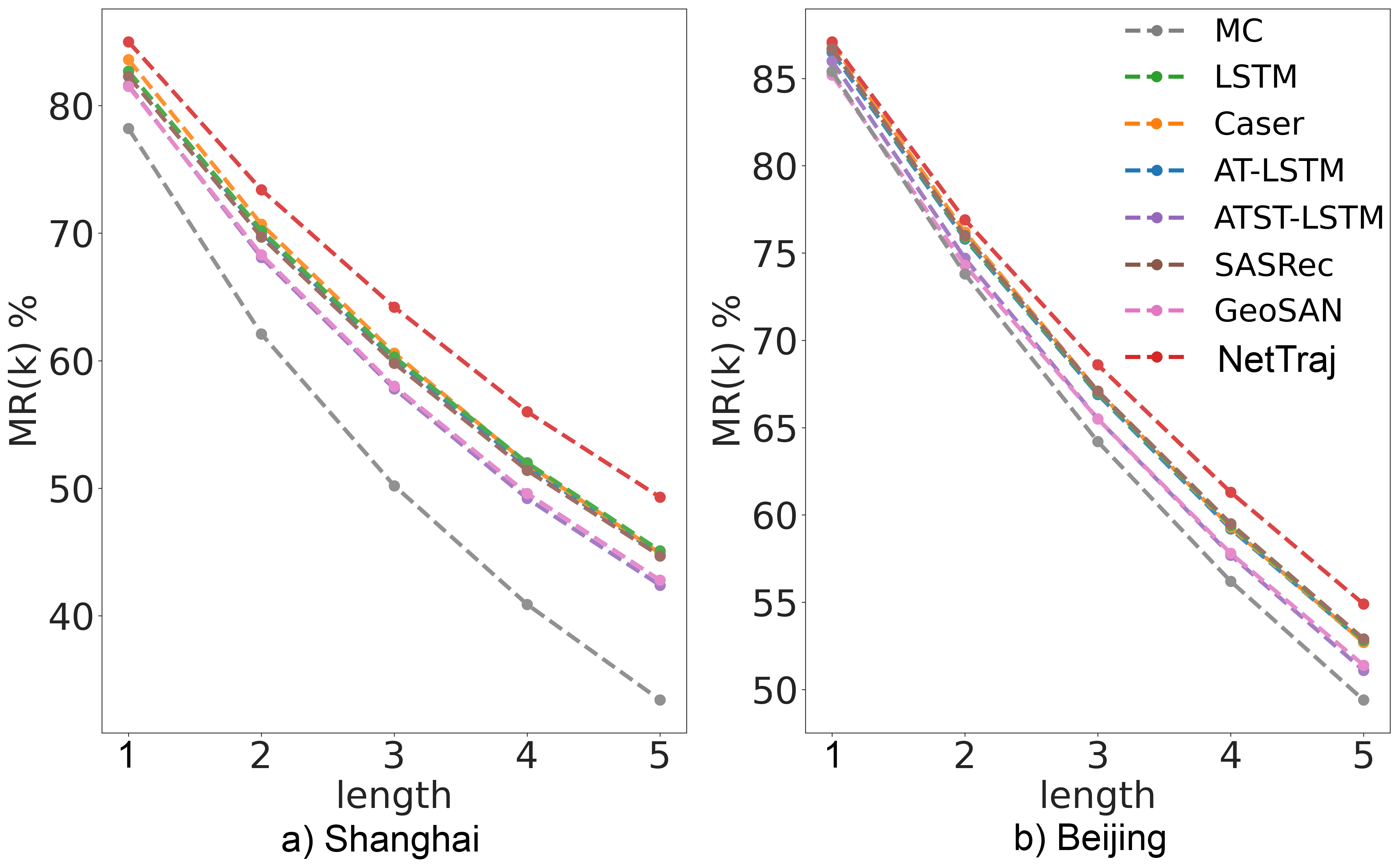}
  \caption{The MR(k) of different trajectory lengths}
  \label{fig:k-MR}
\end{figure}

\begin{figure}[ht]
  \centering
  \includegraphics[width=0.7\linewidth]{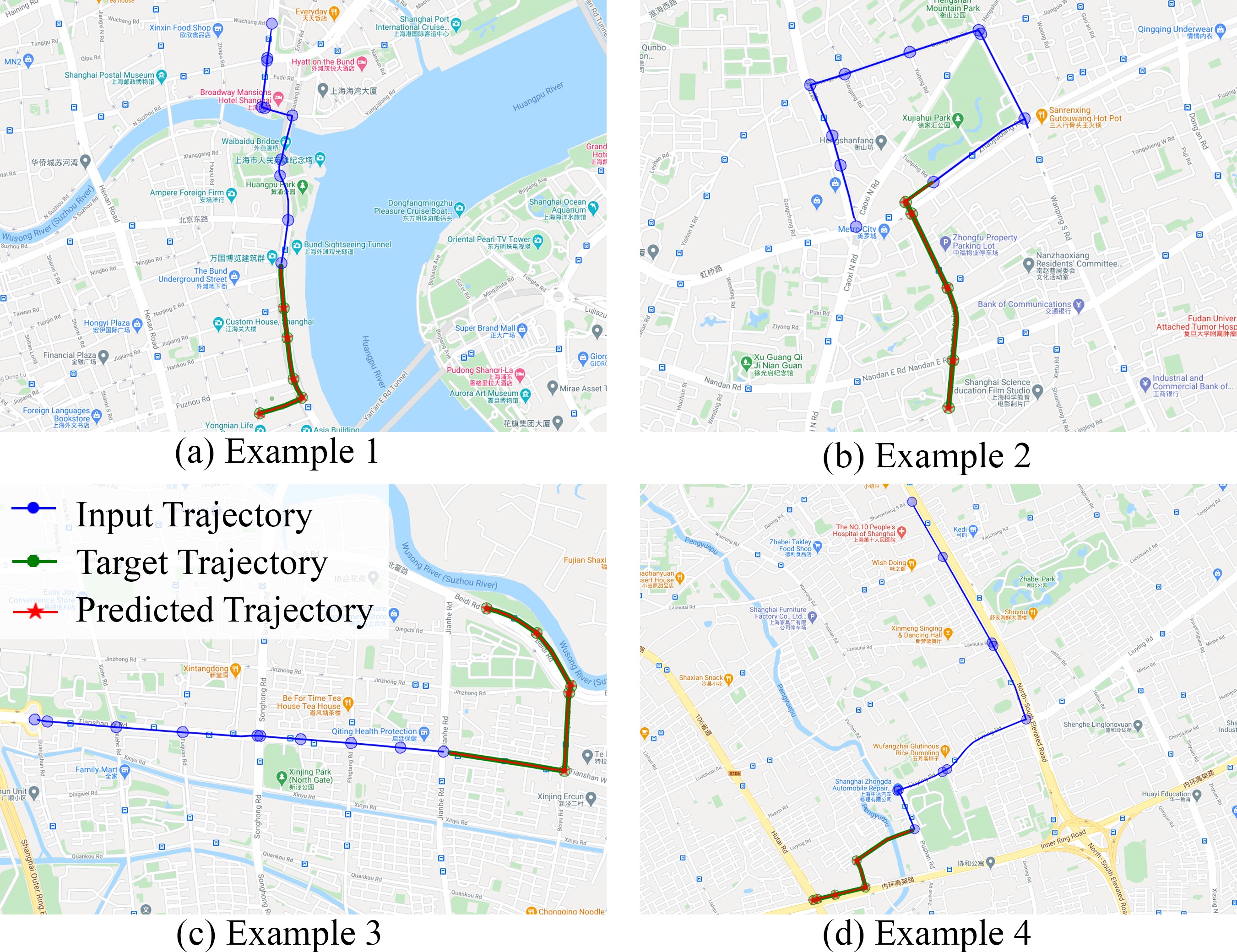}
  \caption{Prediction results in diverse road networks on the Shanghai validation set with NetTraj}
  \label{fig:predict_visualize}
\end{figure}

The results are shown in Table \ref{table:model}. We can observe that for both datasets, the proposed model achieves superior performance than baseline models. In comparison of the state-of-the-art methods, our model shows 8.6\% and 4.8\% reduction in DE for Shanghai dataset and Beijing dataset respectively. This is potentially because existing deep learning methods barely leverage spatial information or capture spatial dependencies at the grid cell level, while our proposed model captures spatial dependencies at road network level with the local graph attention layer. Our model also takes advantage of direction-based trajectory representation that reduces the output dimension and thus simplifies the prediction task. In addition, we adopt a different temporal attention mechanism with a sliding temporal context window to capture both long-term and short-term temporal dependencies. As shown in Fig. \ref{fig:k-MR}, for both Shanghai and Beijing, our model outperforms the baseline methods regarding $MR(k)$, and the gap increases with $k$. This is likely due to error propagation, since the quality of long-range predictions relies on short-range prediction accuracies. Fig. \ref{fig:predict_visualize} shows a few example trajectories from the Shanghai dataset and corresponding predictions based on NetTraj, demonstrating the effectiveness of our proposed model in predicting vehicle trajectories under diverse road network scenarios.
\begin{table*}[ht]
  \centering \footnotesize
  \caption{The performance comparison of different models on Shanghai and Beijing Datasets}
    \begin{tabular}{ccccccc}
    \hline
    \multirow{2}{*}{Model} & \multicolumn{3}{c}{Shanghai} & \multicolumn{3}{c}{Beijing} \\
    & DE (\%) & AMR (\%) & Time (h/ep) & DE (\%) & AMR (\%) & Time (h/ep) \\
    \hline
    MC & 47.0 & 52.8  & - & 34.1 &	65.8 & - \\
    LSTM & 37.5 & 62.2 & 2.15 &	31.6 & 68.3 & 2.85 \\
    Caser & 37.2 & 62.5 & 5.16 & 31.4 &	68.5 & 6.90 \\
    AT-LSTM & 37.7 & 62.0 & 3.63 & 31.7 & 68.1 & 4.82 \\
    ATST-LSTM & 39.7 & 60.0 & 3.48 & 32.8 & 67.0 & 4.45 \\
    SASRec & 38.0 & 61.7 & 1.88	& 31.5 & 68.4 &	2.53 \\
    GeoSAN & 39.5 & 60.2 & 3.34	& 33.0 & 66.9 &	4.39 \\
    \textbf{NetTraj} & \textbf{33.9} & \textbf{65.8} & \textbf{2.75}	& \textbf{30.1} & \textbf{69.7} &	\textbf{3.61} \\
    \hline
    \end{tabular}
  \label{table:model}%
\end{table*}%

Comparing the two datasets, we find that the overall prediction performance in Beijing is consistently better than in Shanghai. The AMR of Beijing dataset is 24.6\% higher than Shanghai data using MC, and 6.4\% - 11.6\% higher than Shanghai using deep learning models. This is potentially because the road network in Beijing is more regularly structured than in Shanghai. As shown in Fig. \ref{fig:dataset}, Beijing has an obvious grid system organizing city circulation, while Shanghai’s road network has more diverse orientations. This may also explain why NetTraj shows higher improvement from baseline methods in the Shanghai data. In complex road networks, NetTraj can significantly improve prediction performance as our proposed direction-based representation can help simplify the road network and the spatial attention mechanism can capture the network structure more efficiently.

Among baseline models, all the deep learning models outperform MC by a notable margin, indicating the effectiveness of deep neural networks in vehicle trajectory prediction. The performance of MC deteriorates rapidly as the prediction length gets longer, suggesting that Markov models perform poorly for long-range trajectory prediction. Compared with Beijing, deep learning models can improve the prediction performance of the Shanghai data more significantly. This indicates that deep learning models are more effective in complex road networks, which is reasonable as neural networks are better at capturing stochastic relationships in complex environments. For both datasets, Caser outperforms the other baseline models regarding $DE$ and $AMR$, demonstrating the effectiveness of convolutional networks in modeling sequential data. Regarding $MR(k)$, Caser performs better than LSTM and SASRec for $MR(1)$, but it does not show superior performance when the prediction length gets longer. This suggests that although convolutional networks are powerful in the next position prediction problem, recurrent and self-attention networks might work better for long-range sequential prediction.

The performance difference between LSTM, AT-LSTM and SASRec is not obvious. LSTM performs slightly better than AT-LSTM, followed by SASRec for the Shanghai data, while in Beijing SASRec slightly outperforms LSTM followed by AT-LSTM. This suggests that neither AT-LSTM nor SASRec is able to show consistent improvement in vehicle trajectory prediction across different road networks. While AT-LSTM and SASRec are capable of handling long-term temporal dependencies, the problem of vehicle trajectory prediction seems to mainly rely on short-term dependencies (e.g., the two previous road segments). This can be further confirmed by Table \ref{table:input_length}, in which we compare the prediction performance of our proposed model with different lengths of input trajectories. It is found that, the AMR can already achieve 62.8\% when $l_{in}=2$, compared to 65.8\% when $l_{in}=10$. This shows that most predictive information can already be captured by the previous two road segments. Table \ref{table:input_length} also shows that the prediction performance gradually improves as $l_{in}$ increases. This indicates that while long-term temporal dependency does not contribute much, it still helps with the prediction performance to some extent. 
\begin{table}[ht!]
  \centering \footnotesize
  \caption{The performance comparison of different input trajectory lengths on Shanghai Dataset using NetTraj}
    \begin{tabular}{cccccccc}
    \hline
    \multirow{3}{*}{Length} & \multicolumn{7}{c}{Shanghai} \\\cline{2-8}
    & DE (\%) & AMR (\%) & \multicolumn{5}{c}{MR(k) (\%)}\\\cline{4-8}
    & & & 1 & 2 & 3 & 4 & 5 \\
    \hline
    2 & 36.9 & 62.8 & 83.4 & 70.8 &	60.8 & 52.4 & 45.9 \\
    4 & 36.3 & 63.4 & 83.9 & 71.5 &	61.9 & 53.3 & 47.1 \\
    6 & 35.3 & 64.4 & 84.4 & 72.2 & 62.7 & 54.3 & 47.9 \\
    8 &	35.0 & 64.7 & 84.6 & 72.4 & 63.0 & 54.7 & 48.2 \\
    10 & 33.9 & 65.8 & 85.0 & 73.4 & 64.2 & 56.0 & 49.3 \\
    \hline
    \end{tabular}
  \label{table:input_length}%
\end{table}%

Among RNN-based models, ATST-LSTM considered spatiotemporal intervals as context infomation, but it does not have better performance over AT-LSTM. This indicates that spatial and temporal intervals between consecutive positions do not contribute much to the vehicle trajectory prediction problem. Among self-attention based models, GeoSAN considers geographic information of intersection nodes using hierarchical griddings, but its performance is even worse than SASRec. This suggests that cell-based representation cannot capture the spatial dependencies in road networks effectively and even has a negative impact on the model performance.

In comparison of the run time of different models, we can find that SASRec has the lowest time cost, followed by LSTM, while Caser requires the longest time for training. This is quite reasonable as self-attention networks skip RNNs and CNNs and enables better parallelization. While AT-LSTM applies only temporal attention and NetTraj considers both temporal and spatial attentions, the time cost of NetTraj is significantly lower than AT-LSTM, which can be explained by the dimension reduction of the output sequences through the direction-based representation. 

\subsection{Ablation Study}
In this section, we conduct extensive ablation studies to quantify the contributions of different components in the proposed model. Since the Shanghai data is more challenging to predict than the Beijing data, we will focus on the Shanghai case here. We denote the based model as NetTraj and drop different components to construct variants, which will be tested and compared against the full model. The components are listed as:
\subsubsection{Direction-based Trajectory Representation (D-TR)}
This denotes the trajectory representation module that transforms a vehicle trajectory to a node sequence and a direction sequence. D-TR allows us to simply predict the direction sequence, which greatly simplifies the prediction problem. With D-TR ablated, both the input and output trajectories are a sequence of intersection nodes.
\subsubsection{Dynamic Spatial Attention (DSA)}
This denotes the spatial attention module used to capture spatial dependencies in road networks. Two models are developed to illustrate the effectiveness of dynamic spatial attention: one replaces DSA with fixed spatial attention module (FSA) in which direction is not incorporated in the computation of spatial attention, and the other drops the spatial attention module.
\subsubsection{Sliding Temporal Attention (STA)}
This denotes the temporal attention module used to capture temporal dependencies in trajectories. Two models are developed to illustrate the effectiveness of sliding temporal attention: one adopts fixed temporal attention module (FTA) in which the temporal attention module aggregates the encoder outputs using attention mechanism, and the other drops the temporal attention module. 
\begin{table*}[ht]
  \centering \footnotesize
  \caption{The performance comparison of different components of NetTraj on Shanghai Dataset}
    \begin{tabular}{ccccccccc}
    \hline
    \multirow{3}{*}{Model} & \multicolumn{8}{c}{Shanghai} \\\cline{2-9}
    & DE (\%) & AMR (\%) & \multicolumn{5}{c}{MR(k)(\%)} & Time (h/ep) \\\cline{4-8}
    & & & 1 & 2 & 3 & 4 & 5 & \\
    \hline
    NetTraj & 33.9 &	65.8 & 85.0 & 73.4 & 64.2 &	56.0 & 49.3 & 2.75 \\
    -D-TR & 37.3 & 62.5 & 83.4 & 70.6 &	60.6 & 52.0 & 45.1 & 4.51 \\
    FSA	& 34.2 & 65.5 &	84.8 & 73.3 & 64.1 & 55.8 & 49.0 & 2.67 \\
    -DSA & 34.6	& 65.1 & 84.7 &	72.8 & 63.5 & 55.2 & 48.5 & 2.32 \\
    FTA	& 34.1 & 65.6 &	85.1 & 73.3 & 64.1 & 55.8 & 48.9 & 2.71 \\
    -DTA & 34.3 & 65.4 & 84.9 &	73.1 & 63.8 & 55.5 & 48.7 & 2.04 \\
    -DSA-DTA & 34.5 & 65.2 & 84.6 &	72.9 & 63.7 & 55.4 & 48.6 & 1.68 \\
    \hline
    \end{tabular}
  \label{table:ablation}%
\end{table*}%

Table \ref{table:ablation} shows the results of the ablation study. We find that the direction-based trajectory representation is crucial for improving both prediction performance and modeling efficiency. Compared with the model without D-TR, the direction-based trajectory representation provides a 9.1\% reduction in DE and 39.1\% reduction in run time. Compared with the baseline LSTM, the model without DSA or DTA can still reduce DE by 8.0\% and improves modeling efficiency by 41.1\%, further validating the effectiveness of our direction-based trajectory representation approach. 

The attention mechanisms are also shown to improve the model performance. 
Compared with the model with no spatial attention at all, fixed spatial attention can lead to a reduction of around 1.2\% in DE, and the integration of directions in spatial attention (or dynamic spatial attention) can further improve the prediction performance slightly. In dynamic spatial attention, the direction information is used to capture the different spatial dependencies for vehicles passing the same intersection from different directions. In addition, compared with the model with no temporal attention considered, our proposed sliding temporal attention results in a reduction of around 1.2\% in DE, superior to the fixed temporal attention with a 0.6\% reduction. This confirms our assumption that the vehicle trajectory prediction problem relies more on short-term temporal dependencies while long-term temporal dependencies only contribute slightly. 

\subsection{Interpretability Study}
To further understand how spatiotemporal dependencies are captured in the NetTraj model, we visualize the attention weights of spatial attention and temporal attention modules. 

\begin{figure}[ht]
  \centering
  \includegraphics[width=0.7\linewidth]{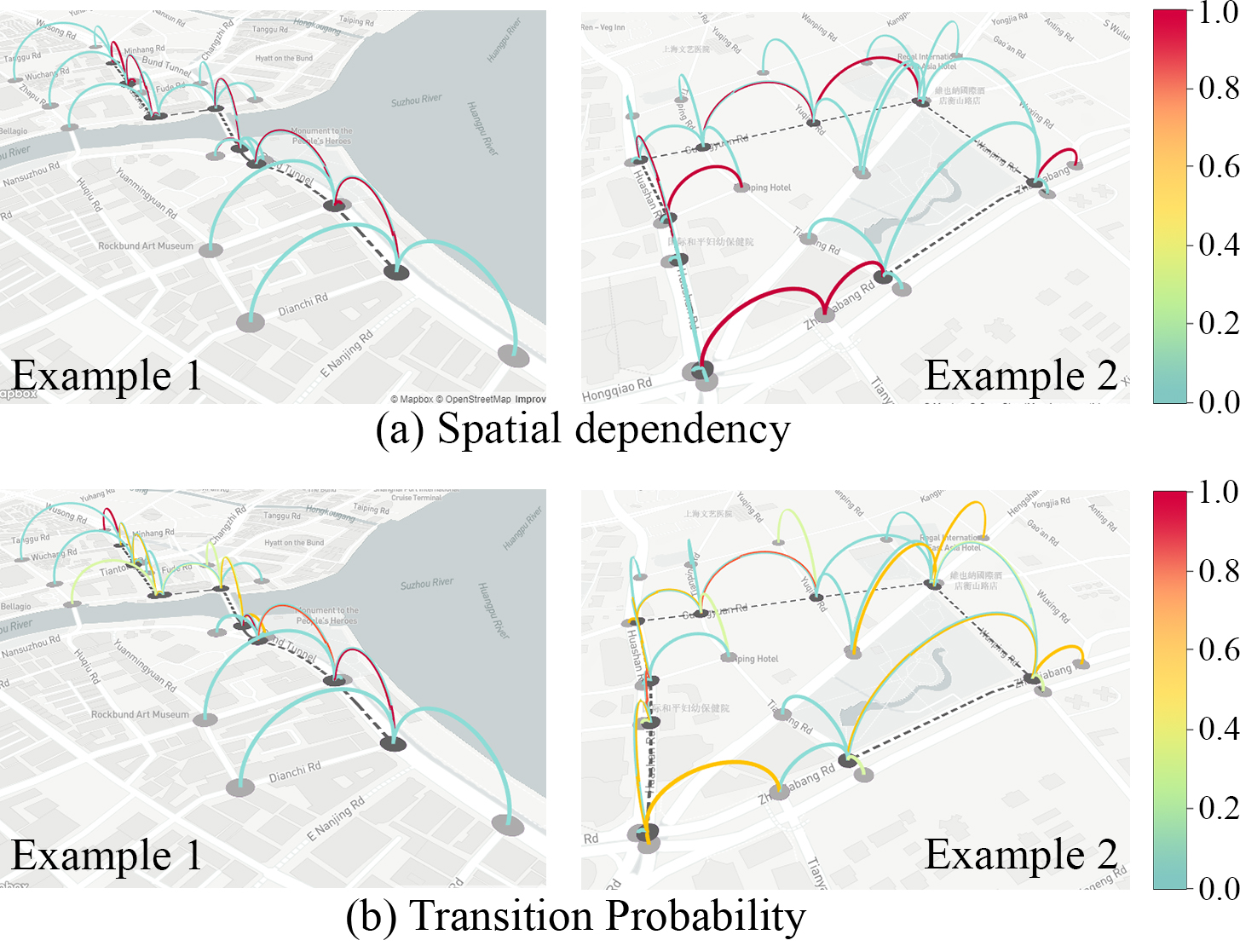}
  \caption{The spatial attention of the input trajectories of the first two examples displayed in Fig. \ref{fig:predict_visualize}. (Dark grey dashed lines represent the input trajectories, dark grey points represent the intersection nodes in trajectories, and light grey points represent the adjacent nodes.)}
  \label{fig:spatial_dependency}
\end{figure}

Fig. \ref{fig:spatial_dependency}a shows the attention weights in spatial attention module for the first two example trajectories displayed in Fig. \ref{fig:predict_visualize}. Recall that the attention weights represent the relative importance of adjacent nodes to a node in a trajectory. Arcs in Fig. \ref{fig:spatial_dependency}a denote the attention weights. Red colors indicate greater weights while blue colors indicate lower weights. It is found that, although each node has multiple adjacent nodes, it usually pays specifically high attention to only one adjacent node. To uncover the mobility patterns underlying in spatial attention weights, we further visualize the transition probabilities between intersection nodes, i.e., the observed probabilities of vehicles moving from a node to another, as shown in Fig. \ref{fig:spatial_dependency}b. Most of the red-highlighted adjacent nodes in Fig. \ref{fig:spatial_dependency}a are also highlighted in Fig. \ref{fig:spatial_dependency}b, indicating that the spatial attention mechanism can to some extent capture the transition probabilities between intersections. However, differences do exist in these two figures, indicating that spatial dependencies of intersections in a road network may vary with driving directions and context features. 

\begin{figure}[ht]
  \centering
  \includegraphics[width=0.7\linewidth]{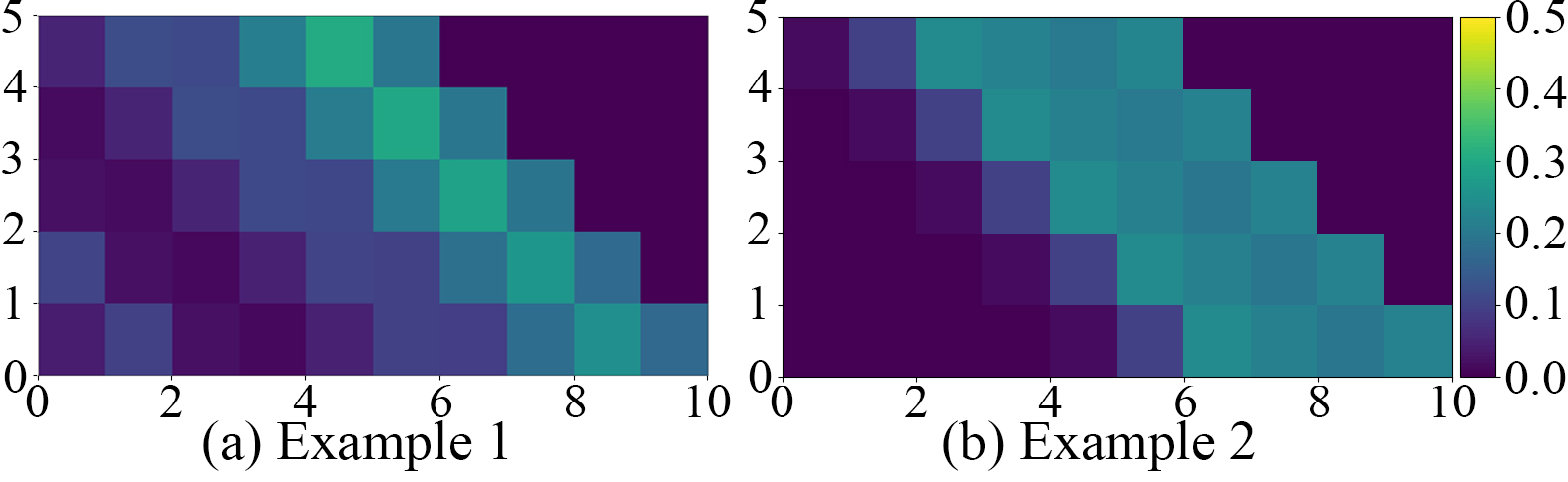}
  \caption{The temporal attention weight matrix of the first two examples displayed in Fig. \ref{fig:predict_visualize} (the x axis represents the position in the sliding temporal window and the y axis represents the output length)}
  \label{fig:temporal_dependency}
\end{figure}

Fig.~\ref{fig:temporal_dependency} shows the attention weights in temporal attention module for the same example trajectories displayed in Fig. \ref{fig:spatial_dependency}. Each heatmap has $l_{out}$ = 5 rows and $l_{in}$ = 10 columns, and each element in the heatmap shows how each road segment (1-10) in the temporal context window influences each predicted road segment (1-5). 
As shown in Fig.~\ref{fig:temporal_dependency}, the temporal attention weight is more evenly distributed than the spatial attention. An interesting observation is that the attention weights on the same diagonal are almost the same. Recall that we adopt a sliding context window for temporal attention computation. This indicates that the weighted influence of the same road segment in the input trajectory to different road segments in the output trajectory is mostly the same. We can also find that, in the input trajectory, road segments that are closer to the output trajectory generally have higher attention weights, validating our assumption that trajectory prediction mainly relies on short-term dependency. 

\section{Conclusion}\label{Conclusion}
Trajectory prediction of vehicles at the city scale is of great importance to various location-based applications such as vehicle navigation, traffic management, and location-based recommendations. In this paper, we propose a novel seq2seq model with directional representation and spatiotemporal attention mechanisms, named NetTraj, for vehicle trajectory prediction in city-scale road networks. Unlike existing methods that represent trajectories as sequences of grid cells, road segments or driver intentions, we propose a direction-based trajectory representation approach which represents each trajectory as a sequence of intersections and associated movement directions. This new representation greatly simplifies the trajectory prediction problem and improves both the prediction accuracy and modeling efficiency. To capture the dynamic spatial dependencies in road networks, a local graph attention module is introduced, in which a small graph is formed over neighborhood nodes for each intersection in the trajectory and an attention mechanism is applied to calculate the weighted influence between neighborhood nodes. Finally, a LSTM encoder-decoder network with a sliding temporal attention mechanism is adopted to capture both long-term and short-term temporal dependencies. Experiments on two real-world taxi trajectory datasets show that NetTraj can achieve superior prediction performance than state-of-the-art algorithms.

There are several directions for future research. First, some specific physical attributes of road segments are not considered in our model due to lack of data, and road attributes such as type of road, speed limit, real-time traffic may also affect vehicle driving behaviors. Second, our proposed trajectory representation approach is developed for 2D road networks while future studies can extend it to 3D network structure, such as 3D pedestrian networks with overpasses and underground tunnels. This is particularly useful for cities like Hong Kong with 3D transportation networks. Third, existing studies mostly focus on trajectory prediction for vehicles en route to a destination. However, sometimes vehicles do not have a clear or fixed destination, e.g., cars cruising for parking space or taxis cruising for passengers. The routing patterns of cruising vehicles are quite different, as their goal is no longer to find the fastest path to some places. Trajectory prediction of cruising vehicles can provide trajectory-level cruising strategies, improve the efficiency of urban traffic systems and reduce the number of vehicles on the road.

\section*{Acknowledgments}
The computations were performed using research computing facilities offered by Information Technology Services, the University of Hong Kong.

\bibliographystyle{unsrt}  
\bibliography{ref}

\end{document}